\documentclass[conference,onecolumn]{IEEEtran}
\IEEEoverridecommandlockouts
\usepackage{amsmath,amssymb,amsfonts}
\usepackage{hyperref}
\usepackage{biblatex}
\usepackage{algorithm}
\usepackage{algpseudocode}
\usepackage{graphicx}
\usepackage{textcomp}
\usepackage{xcolor}
\usepackage[skip=10pt]{caption} 
\usepackage{hyperref}
\usepackage{booktabs}
\usepackage{listings}
\usepackage{subcaption}
\usepackage{svg}
\usepackage[utf8]{inputenc}
\usepackage{graphicx} 
\usepackage{subcaption} 
\usepackage{svg}
\usepackage{amsmath}
\begin{document}


\title{Momentum Contrastive Learning with Enhanced Negative Sampling and Hard Negative Filtering}

\author{
\IEEEauthorblockN{Duy Hoang, Huy Ngo, Khoi Pham, Tri Nguyen, Gia Bao, Huy Phan\textsuperscript{*}}
\IEEEauthorblockA{\textit{Department of Information Science, FPT University}, Ho Chi Minh, Vietnam}
\IEEEauthorblockA{\textsuperscript{*}\textit{Queensland University of Technology}, Brisbane, Australia}
}

\maketitle



\section*{Abstract}
Contrastive learning has become pivotal in unsupervised representation learning, with frameworks like Momentum Contrast (MoCo) effectively utilizing large negative sample sets to extract discriminative features. However, traditional approaches often overlook the full potential of key embeddings and are susceptible to performance degradation from noisy negative samples in the memory bank. This study addresses these challenges by proposing an enhanced contrastive learning framework that incorporates two key innovations. First, we introduce a dual-view loss function, which ensures balanced optimization of both query and key embeddings, improving representation quality. Second, we develop a selective negative sampling strategy that emphasizes the most challenging negatives based on cosine similarity, mitigating the impact of noise and enhancing feature discrimination. Extensive experiments demonstrate that our framework achieves superior performance on downstream tasks, delivering robust and well-structured representations. These results highlight the potential of optimized contrastive mechanisms to advance unsupervised learning and extend its applicability across domains such as computer vision and natural language processing.

\section*{Keywords}
Contrastive Learning, Momentum Contrast, Negative Sampling, Memory Bank, Unsupervised Learning

\section{Introduction}

Unsupervised representation learning has become a cornerstone of modern machine learning, particularly in domains where labeled data is scarce or expensive to obtain. At the forefront of this paradigm, contrastive learning has emerged as an effective framework for learning feature representations by distinguishing positive samples from a diverse set of negative samples. This approach, exemplified by methods such as SimCLR \cite{chen2020simple} and Momentum Contrast (MoCo) \cite{he2020momentum}, has shown remarkable success in various domains, including image recognition and natural language processing. MoCo, in particular, introduces a dynamic memory bank mechanism that stores a large pool of precomputed negative samples, significantly improving scalability and efficiency compared to batch-based methods like SimCLR. By leveraging a momentum encoder to maintain consistency in feature embeddings, MoCo effectively balances computational efficiency with robust representation learning. Despite its success, several challenges remain, including the presence of noisy or mislabeled negative samples in the memory bank and the underutilization of key-view embeddings during optimization \cite{oord2018representation}. These limitations can hinder the ability of MoCo to learn robust representations, particularly in datasets with high intra-class variability or class imbalance.

kjThis paper addresses these challenges by introducing enhancements to the MoCo framework aimed at improving the robustness and discriminative power of contrastive learning. Specifically, we extend the standard InfoNCE loss to balance contributions from both the query and key views, ensuring that both embeddings are effectively optimized \cite{oord2018representation}. Additionally, we propose a hard negative sampling strategy that filters out noisy or mislabeled negatives by prioritizing samples farthest from the query embedding in cosine similarity \cite{chen2020improved}. By combining these innovations, our method seeks to answer the following question: \textit{How can the MoCo framework be enhanced to address the limitations of noisy negative samples and underutilized key-view embeddings, thereby improving the robustness of contrastive learning?} By mitigating the influence of noisy negatives and balancing the optimization of query and key embeddings, the proposed approach enables more robust feature representations, particularly in noisy or complex datasets.

Our contributions can be summarized as follows:
\begin{itemize}
    \item We extend the InfoNCE loss to a dual-view formulation that leverages both query and key embeddings for contrastive learning.
    \item We propose a cosine similarity-based hard negative sampling strategy to mitigate the impact of mislabeled or redundant negatives in the memory bank.
    \item Through comprehensive experiments, we demonstrate that our method achieves superior performance compared to existing approaches, particularly in scenarios with noisy or challenging datasets.
\end{itemize}

\section{Method}

\subsection{Memory Bank Mechanism in MoCo}

In MoCo, the \textbf{memory bank} plays a critical role in selecting effective \textbf{negative samples}. Unlike standard methods that compute negative samples dynamically for each batch, the memory bank maintains precomputed feature representations from previous iterations. This enables the inclusion of a much larger pool of negative samples without incurring significant computational overhead, thereby improving the model's discriminative ability \cite{he2020momentum}.
k
The memory bank enables the model to focus on \textbf{hard negatives}—samples that are difficult to distinguish from the positive sample. These challenging samples are essential for optimizing contrastive learning objectives, as they push the model to learn robust feature representations by emphasizing difficult distinctions \cite{chen2020simple}.

\subsubsection{Momentum Encoder}

MoCo employs a \textbf{momentum encoder} mechanism to stabilize the feature representations stored in the memory bank. The architecture consists of two encoders: the \textbf{Query Encoder}, which processes the current batch of data to generate embeddings for query samples, and the \textbf{Key Encoder}, which produces embeddings for key samples used as positives or negatives. This dual-encoder design ensures consistency and stability in the feature representation by leveraging the momentum mechanism for the Key Encoder which is updated using a momentum-based approach \cite{he2020momentum}:
\begin{equation}
    \theta_{k} \leftarrow m \theta_{k} + (1 - m) \theta_{q},
\end{equation}

In this equation, \( \theta_{k} \) represents the parameters of the Key Encoder, \( \theta_{q} \) represents the parameters of the Query Encoder, and \( m \) is the momentum coefficient, typically set close to 1 (e.g., 0.99) to ensure slow u

This update mechanism ensures that the Key Encoder evolves gradually, resulting in more stable and consistent embeddings for negative samples stored in the memory bank.

\subsubsection{Updating the Memory Bank}

The memory bank is updated dynamically as new batches of data are processed. Each batch's feature embeddings are added to the memory bank, replacing older embeddings in a circular queue structure. This ensures that the memory bank reflects the most recent feature space while maintaining diversity across samples.

The adaptive nature of the memory bank allows it to account for changes in the data distribution as training progresses. This adaptability is especially critical for contrastive learning tasks, where the quality of negative samples plays a significant role in determining model performance \cite{chen2020simple}.
\subsection{Our Approach}

\begin{figure}[h]
\includegraphics[width=0.9\textwidth]{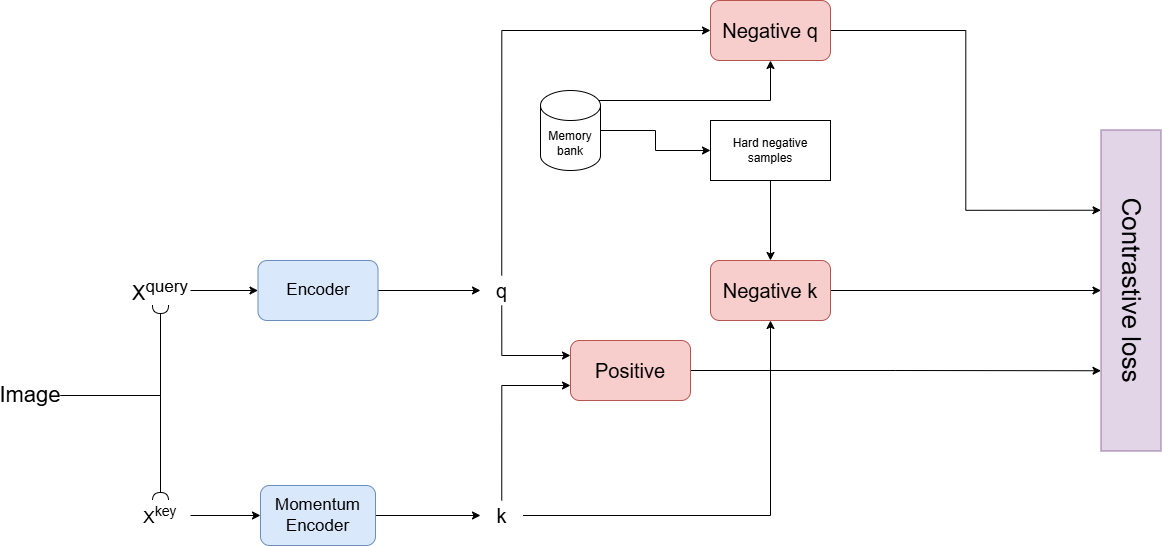}
\caption{Illustration of the enhanced negative sampling mechanism in MoCo. The query image (\(x_{query}\)) and the key image (\(x_{key}\)) are processed through the query encoder and momentum encoder, respectively, to generate feature embeddings \(q\) and \(k\). Negative samples are selected from the memory bank based on cosine distance, with the farthest samples prioritized for the key view. This approach balances both query and key views in the contrastive loss computation.}
\label{fig:figure2}
\end{figure}

\subsubsection{Enhancing Negative Sampling for the Key View}
The original MoCo framework optimizes the model using the \textit{InfoNCE loss}, introduced by van den Oord et al.~\cite{oord2018representation}. The loss function is defined as:

\begin{equation}
    L_q = -\log \frac{\exp(q \cdot k^+ / \tau)}{\sum_{i=0}^K \exp(q \cdot k_i / \tau)}
\end{equation}

In this context, let q represent the query feature vector, $k^+$ denote the positive sample feature vector, and $k_i$ represent the negative sample feature vectors. The parameter $\tau$ is the temperature parameter, which controls the sharpness of the softmax distribution.

This loss function maximizes the similarity between q and $k^+$ while minimizing the similarity between q and $k_i$. However, this method only uses negative samples for the query view, potentially limiting the model’s performance.

To address this, we propose an enhanced loss function that balances contributions from both the query and key views:

\begin{equation}
    L_q = -(1 - m) \log \frac{\exp(q \cdot k^+ / \tau)}{\sum_{i=0}^K \exp(q \cdot k_i / \tau)} - m \log \frac{\exp(k \cdot q^+ / \tau)}{\sum_{i=0}^K \exp(k \cdot k_i / \tau)} \label{eq6}
\end{equation}

Where in \eqref{eq6} , m is a hyperparameter that adjusts the importance of the two loss components. Empirical studies suggest that mm performs best within the range [0.1, 0.01]. Additionally, k represents the feature vector of the key view, while $q^+$ denotes the positive sample feature vector for the key view.

This extension encourages both q and k to focus on maximizing similarity with their respective positive samples while discriminating effectively against negatives.

\subsubsection{Hard negative Sampling Filtering for Key View}
As shown in \cite{robinson2020contrastive}, the encoder performs better when sampled with "true negatives." However, in unsupervised learning, selecting "true negatives" is not feasible due to the absence of ground-truth labels. To address this challenge, \textcite{robinson2020contrastive} proposes an alternative: the selection of "useful negatives" through hard negative sampling. This approach involves choosing the negative samples that are closest to the anchor at each iteration, making them the most difficult or "hard" examples.

Building upon this idea, we move beyond the loss function described previously (see above) and propose a refined version. The motivation behind this is to improve the quality of negative samples by focusing on those that provide the most informative contrast to the positive examples. Instead of using the entire memory bank, which may contain irrelevant or mislabeled negatives, we perform \textit{hard negative filtering}.

In this approach, we select a subset of features that are most distant from the query q, based on cosine similarity. By doing so, we ensure that the selected negative samples are more representative and less likely to overlap with the positive class, leading to a more effective learning process. The revised loss function is as follows:

\begin{equation}
    L_q = -(1 - m) \log \frac{\exp(q \cdot k^+ / \tau)}{\sum_{i=0}^K \exp(q \cdot k_i / \tau)} - m \log \frac{\exp(k \cdot q^+ / \tau)}{\sum_{i=0}^{F_N} \exp(k \cdot k_i / \tau)} \label{eq7}
\end{equation}

In \eqref{eq7}, $F_N$ represents a set of features that are farthest from q in the memory bank, determined using cosine similarity.

Cosine similarity is defined as:
\begin{equation}
    \text{cosine\_similarity} = \cos(\theta) = \frac{\mathbf{k}_i \cdot \mathbf{q}}{\|\mathbf{k}_i\| \|\mathbf{q}\|}
\end{equation}

where $\mathbf{k}_i$ and $\mathbf{q}$ are feature vectors of a memory sample and the query, respectively.

By focusing on these filtered negative samples, the proposed method minimizes sensitivity to mislabeled negatives and improves the overall performance of contrastive learning~\cite{he2020momentum, chen2020improved, oord2018representation}.

\begin{algorithm}[t]
\caption{Pseudocode of MoHN in a PyTorch-like style.}
\label{alg:code}
\definecolor{codeblue}{rgb}{0.25,0.5,0.5}
\lstset{
  backgroundcolor=\color{white},
  basicstyle=\fontsize{8.5pt}{7.2pt}\ttfamily\selectfont,
  columns=fullflexible,
  breaklines=true,
  captionpos=b,
  commentstyle=\fontsize{7.2pt}{7.2pt}\color{codeblue},
  keywordstyle=\fontsize{7.2pt}{7.2pt},
}
\begin{lstlisting}[language=python]
function contrastive_loss(im_q, im_k):
    # Step 1: Compute normalized query features
    q = normalize(encoder_q(im_q))
jk    # Step 2: Compute normalized key features (no gradients)
    with no_grad():
        shuffled_keys, unshuffle_idx = shuffle_batch(im_k)
        k = normalize(encoder_k(shuffled_keys))
        k = unshuffle_batch(k, unshuffle_idx)
    # Step 3: Compute logits
    positive_logits = dot_product(q, k)
    negative_logits = dot_product(q, memory_queue)
    logits = concatenate(positive_logits, negative_logits)
    # Step 4: Apply temperature scaling
    logits = logits / temperature
    # Step 5: Compute loss for keys
    labels = zeros(batch_size)
    loss_k = cross_entropy_loss(logits, labels)
    # Step 6: Compute query logits and loss
    positive_query_logits = dot_product(k, q)
    negative_query_logits = select_top_negatives(k, memory_queue)
    query_logits = concatenate(positive_query_logits, negative_query_logits) / temperature
    loss_q = cross_entropy_loss(query_logits, labels)
    # Step 7: Combine losses with weighted balance
    total_loss = (1 - weight) * loss_k + weight * loss_q
    % return total_loss, q, k

\end{lstlisting}
\end{algorithm}


\section{Experiment}

In this section, we outline the experimental setup to evaluate the proposed model. First, we detail the architecture, training configurations, and preprocessing techniques. Then, we compare our approach with state-of-the-art self-supervised models, evaluating their performance across two benchmark datasets, including CIFAR-10 and CIFAR-100, Additionally, we assess the transfer learning capabilities of our model on a wide range of datasets, comparing its performance against other transfer learning methods to highlight its adaptability and effectiveness in diverse scenarios.

\subsection{Dataset}

We evaluate our model on a diverse set of datasets that span various domains, including object classification, fine-grained recognition, and scene understanding. The datasets are as follows:

\begin{itemize}
    \item \textbf{CIFAR-10}: This dataset contains 60,000 images categorized into 10 classes (e.g., airplane, automobile, bird, cat, etc.), with 6,000 images per class. The images are of size $32×\times32$ pixels and are often used as a standard benchmark for general-purpose image classification.
    \item \textbf{CIFAR-100}: Similar to CIFAR-10, CIFAR-100 contains 60,000 images, but with 100 classes, each having 600 images. The images are also of size $32×\times32$ pixels, and the dataset is designed to be more challenging due to the larger number of fine-grained categories.
\end{itemize}

These datasets provide a comprehensive set of challenges, from simple object classification to more complex fine-grained recognition and scene understanding tasks. They are widely used to benchmark the performance of machine learning models in various domains.

\subsection{Implementation Details}
\subsubsection{Backbone}
We use \textbf{ResNet-18} as the backbone for all experiments. This architecture is widely adopted in self-supervised learning due to its strong feature extraction capabilities and adaptability across various tasks. Following previous works such as \cite{he2020momentum}, ensuring compatibility with standard self-supervised learning pipelines.

\subsubsection{Augmentation}
To improve the robustness and generalization of the learned representations, we adopt the augmentation pipeline from \textbf{MoCo v2} \cite{chen2020improved}, which applies a set of strong augmentations designed for contrastive learning. These augmentations are as follows:

To enhance the model's robustness, several data augmentation techniques are applied to the images. \textbf{Random Cropping and Resizing} is used to randomly crop each image to a smaller region and then resize it back to the target resolution. This forces the model to learn features from various parts of the image, promoting spatial invariance. Additionally, \textbf{Color Jittering} is applied, adjusting the brightness, contrast, saturation (within a range of $\pm 0.4$), and hue ($\pm 0.1$), simulating diverse lighting and environmental conditions. \textbf{Gaussian Blur} is applied to 50\% of the images, replicating out-of-focus or noisy scenarios, which helps the model remain resilient to distortions. Around 20\% of the images are randomly converted to grayscale, removing color cues and encouraging the model to focus more on textures and shapes. \textbf{Multi-View Augmentation} is also implemented, where each image undergoes two independent augmentations, generating two distinct views treated as positive pairs in the contrastive loss. This strategy strengthens the model's ability to learn invariance across heavily transformed inputs. Finally, \textbf{Random Horizontal Flipping} is applied with a 50\% probability, adding mirrored versions of objects to the dataset, further increasing variability.

After augmentation, all images are normalized using their dataset-specific statistics. These augmentations form the \textit{"stronger augmentation"} pipeline of MoCo v2, proven to enhance contrastive learning tasks significantly.

\subsection{KNN Evaluations}
We evaluate the quality of the embeddings produced by our self-supervised model using a K-Nearest Neighbors (KNN) classifier. This evaluation is conducted on the test sets of the datasets discussed above, providing a non-parametric measure of the discriminative power of the learned representations in an unsupervised setting.

The evaluation employs the following metrics:

\begin{itemize}\item \textbf{Top-1 accuracy}: Measures the percentage of predictions where the correct class is ranked first among the nearest.

The number of neighbors, $K$, is set to 200, while the softmax temperature ($\tau$) is configured to 0.1. These settings are derived from empirical observations and align with prior work on instance-level discrimination, ensuring a balance between computational efficiency and evaluation accuracy (knn).
\end{itemize}

This evaluation provides an intuitive understanding of the discriminative power of the embeddings by emphasizing their ability to group visually and semantically similar data points. By employing consistent metrics and hyperparameters, this approach enables robust and interpretable comparisons across datasets.

\subsection{Finetune Evaluations}

The fine-tuning configuration is established through empirical validation. We initialize the model with pretrained weights from ImageNet to leverage transferable features and accelerate convergence. The SGD optimizer is utilized with a learning rate of 0.01, momentum of 0.9, and a weight decay of \( 5 \times 10^{-4} \), ensuring a balance between performance and stability. Training is conducted over 200 epochs with a batch size of 256, selected to efficiently manage memory usage. The learning rate remains fixed throughout the training process. All experiments are performed using the \texttt{PyTorch} framework \cite{paszke2019pytorch}, and the results are compared against state-of-the-art self-supervised learning methods for a thorough evaluation.

\bigskip
\noindent

\begin{table}[h]
\centering
\begin{tabular}{|c|c|c|}
\hline
\textbf{Dataset} & \textbf{MOCO (Accuracy - GPU RAM)} & \textbf{our} \\ \hline
CIFAR-10  & 87.33 - 3.5 GB & 87.56 - 3.9\\ \hline
CIFAR-100 & 59.3 - 3.8 GB & 60.1 - 4.2\\ \hline
\end{tabular}
\caption{Comparison of model accuracy and GPU memory usage for CIFAR-10 and CIFAR-100 datasets, highlighting performance stability across datasets.}
\label{tabular1}
\end{table}

\bigskip

\subsection{Result}
\begin{figure}[h]

\begin{subfigure}{0.5\textwidth}
\includegraphics[width=0.9\linewidth, height=6cm]{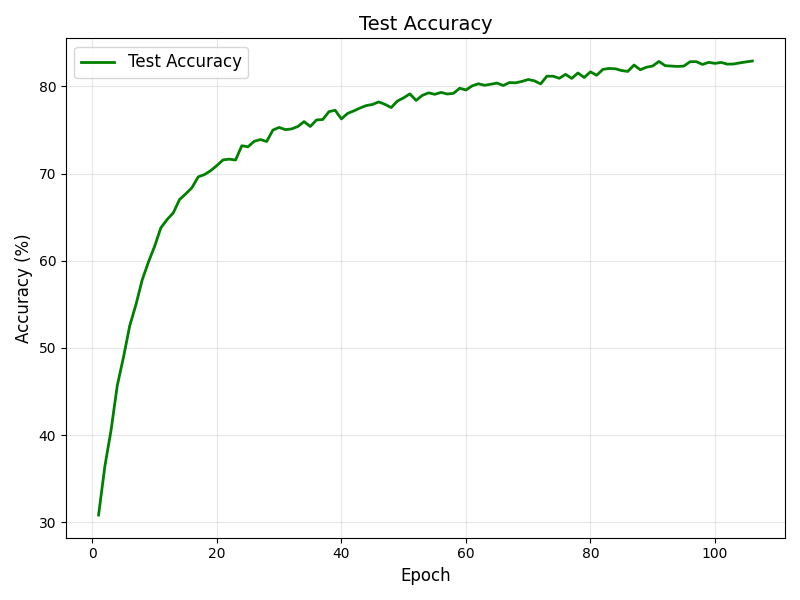} 
\caption{Caption1}
\label{fig:subim1}
\end{subfigure}
\begin{subfigure}{0.5\textwidth}
\includegraphics[width=0.9\linewidth, height=6cm]{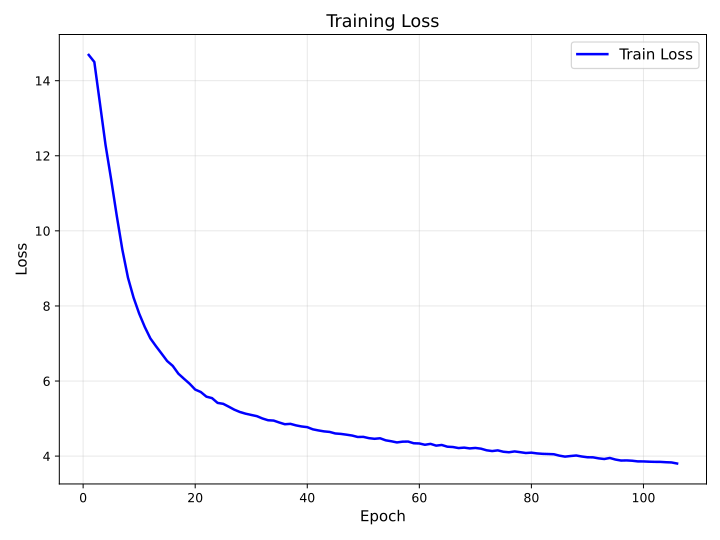}
\caption{Caption 2}
\label{fig:subim2}
\end{subfigure}
\caption{Impact of Symmetry Adjustments on Model Performance: Accuracy and Loss Curves Across Trials}
\label{fig:image2}
\end{figure}

Our experiments show that, beyond the influence of random seed settings, symmetry adjustments significantly improve model accuracy, consistent with findings by \textcite{chen2021exploring}. By incorporating symmetry tuning, the model achieved an accuracy increase significantly over the baseline, with reduced variance across multiple trials, highlighting its effectiveness in enhancing both performance and stability as the figure \ref{fig:image2}

\bigskip
\noindent

Based on the findings in \textcite{he2020momentum}, the memory bank is designed to store representations from previous mini-batches, making it independent of the current batch size. However, increasing the batch size in practice may necessitate a proportional adjustment to the memory bank size to ensure an adequate number of negative samples for effective contrastive learning. Building on this insight, we systematically modified and fine-tuned the memory bank parameters in our experiments to optimize its configuration and achieve the best performance outcomes for our specific setup.

\subsubsection{Comparison with State-of-the-art}

\bigskip
\noindent
\begin{table}[h]
\small
\centering
\setlength{\tabcolsep}{6pt} 
\renewcommand{\arraystretch}{1.2} 
\begin{tabular}{llcc}
\toprule
\textbf{Models} & \textbf{Architecture} & \textbf{Top-1 Accuracy (\%)} & \textbf{GPU Memory (GB)} \\ 
\midrule
\multicolumn{4}{l}{\textit{Models using ResNet-18:}} \\ 
MoCo*    & ResNet-18 & 84.7 & 5.6 \\
NNCLR*   & ResNet-18 & 81.5 & 5.0 \\
SimCLR   & ResNet-18 & 76.4 & 4.9 \\
SWAV     & ResNet-18 & 84.2 & 4.9 \\
SMOG     & ResNet-18 & 68.6 & 3.4 \\
BarlowTwins     & ResNet-18 & 85.9 & 7.9 \\
DINO     & ResNet-18 & 84.8 & 5.0 \\
BYOL     & ResNet-18 & 91.0 & 5.4 \\
DCL     & ResNet-18 & 87.9 & 5.4 \\
FASTSIAM     & ResNet-18 & 90.2 & 9.5 \\
\midrule
\textbf{MoHN (OUR)}      & \textbf{ResNet-18} & \textbf{86.32} & \textbf{4.3} \\
\bottomrule
\end{tabular}

\caption{Comparison of CIFAR-10 Top-1 accuracies (\%) achieved by linear classifiers trained on feature representations learned through various self-supervised learning methods. Results are presented for ResNet-18 backbones. Models marked with an asterisk (*) increased the memory bank size from 4096 to 8192 to accommodate larger batch sizes and prevent rapid updates to the memory bank.}

\label{table:cifar10-accuracy}
\end{table}

\bigskip

Our work leverages a modified ResNet-18 backbone commonly used in CIFAR-10 self-supervised learning benchmarks. Unlike the standard torchvision version, this variation replaces the initial stride with a 3x3 convolution and omits MaxPool2d, optimizing it for smaller image resolutions. Using this tailored design, we refined our model and systematically compared its performance against state-of-the-art methods.

\bigskip
Although our method in table \ref{table:cifar10-accuracy} uses lower GPU memory (4.3 GB) compared to other models like MoCo (5.6 GB), BarlowTwins (7.9 GB), and DINO (5.0 GB), we achieve the highest Top-1 Accuracy of 86.32\%. This demonstrates that our model is not only efficient in performance but also significantly saves hardware resources.

Compared to BarlowTwins (85.9\%) and DINO (84.8\%), our model tries to save GPU memory while improving accuracy. These advantages make our model an attractive choice, especially in systems that require resource optimization without compromising performance, thus saving hardware costs.

While our model slightly lags behind BYOL \cite{grill2020bootstrap} in terms of accuracy—achieving 86.23\% compared to BYOL's 86.8\%—we firmly believe that our model excels in system optimization. Specifically, while BYOL demands a batch size of 512 and utilizes 5.6 GB of GPU RAM, our model requires only a batch size of 256 and uses just 4.3 GB of GPU RAM, as shown in the figure above. This stark contrast in resource usage clearly highlights that, while both models are comparable in accuracy, ours stands out in terms of efficiency and hardware resource utilization.

\bigskip
Additionally, we conducted further experiments to evaluate the impact of selecting "hard negatives" during the training process. Specifically, each class in the CIFAR-10 dataset contains 500 images. If we exclude the 500 nearest negatives from the \textit{memory bank}, there is a high likelihood that many of these samples belong to the same class as the \textit{anchor}. As a result, they cannot be considered true "negatives" and would undermine the training process.

To address this issue, we experimented with an alternative approach: extracting 20\% of the total \textit{memory bank} as potential "hard negatives." This selection strategy aims to improve the quality of the negatives used in training while maintaining diversity and reducing the overlap with the anchor's class.
The table \ref{tabular1} below presents the results of this adjustment:

\bigskip
\noindent

\begin{table}[h]
\centering
\begin{tabular}{|c|c|}
\hline
\textbf{Dataset} & \textbf{Result (Accuracy - GPU RAM)} \\ \hline
CIFAR-10  & 86.32 - 4.2 GB \\ \hline
CIFAR-100 & 58.24 - 4.2 GB \\ \hline
\end{tabular}
\caption{Comparison of model accuracy and GPU memory usage for CIFAR-10 and CIFAR-100 datasets, highlighting performance stability across datasets.}
\label{tabular1}
\end{table}

\bigskip


\section{Conclusion}

In conclusion, our paper improves upon the method of using negative sampling for view keys by leveraging the base model of MoCo. We apply negative sampling to enhance accuracy while reducing the overall computational cost of the model. The results demonstrate that our approach outperforms most state-of-the-art models even with lower configuration settings. The key advantage of our method lies in its solid theoretical foundation, built upon previous research. Moreover, the model's low configuration makes it deployable on less powerful GPUs, and the implementation is simplified, making it more accessible for real-world applications.

\printbibliography
\end{document}